# The Go Transformer: Natural Language Modeling for Game Play

Matthew Ciolino
PeopleTec Inc.
Huntsville, AL, USA
matt.ciolino@peopletec.com

Josh Kalin
Department of Computer Science and
Software Engineering
Auburn University
Auburn, AL, USA
jzk0098@auburn.edu

David Noever
PeopleTec Inc.
Huntsville, AL, USA
david.noever@peopletec.com

**This work applies natural language modeling to generate plausible strategic moves in the ancient game of Go. We train the Generative Pretrained Transformer (GPT-2) to mimic the style of Go champions as archived in Smart Game Format (SGF), which offers a text description of move sequences. The trained model further generates valid but previously unseen strategies for Go. Because GPT-2 preserves punctuation and spacing, the raw output of the text generator provides inputs to game visualization and creative patterns, such as the Sabaki project's game engine using auto-replays. Results demonstrate that language modeling can capture both the sequencing format of championship Go games and their strategic formations. Compared to random game boards, the GPT-2 fine-tuning shows efficient opening move sequences favoring corner play over less advantageous center and side play. Game generation as a language modeling task offers novel approaches to more than 40 other board games where historical text annotation provides training data (e.g., Amazons & Connect 4/6)**

I. INTRODUCTION

This paper explores the application of language transformers to play Go, or at least to generate plausible Go game boards. The motivation stems in part from the potential advantages of learning advanced strategies and for providing additional feature-rich inputs for training and generating algorithms both in language, imagery, audio and game-playing. The original contributions include the fine-tuning methods for applying pre-trained language transformers to generate abstract game notations and the visualization of that output as dueling, personified game players. It is worth noting that comparing statistical metrics for language is not without its ambiguities and controversies, but a game victory offers concrete scoring and defines a winner between different players or potentially future language models doing other tasks.

The 2500-year old board game of Go attracts strategy play by 450 million participants, mainly from Asia [1]. For the machine learning community, the game holds special attraction as enumerating all possible moves exceeds the iconic benchmark of all combinatorial calculators, namely more Go moves than the number of atoms in the universe [2, 3]. As played with black and white stones on a 19x19 grid (Figure 1), the game itself features both the walling in of your domain or territory (open grid spaces), and the capturing of your opponent's stones while extending your territory. A point system determines the winner based on number of pieces on the board and open spaces walled-in to your territory. Because computer players cannot enumerate all possible moves, human ingenuity and imagination has dominated traditional brute force programming. Unlike the best computer programs for other board games, the best Go programs have traditionally begun play only at the level of an advanced beginner [4]. This difficulty contrasts with the machine superiority for backgammon, chess, Scrabble, Chinese and western checkers, Connect-4/6 and Mancala.

By applying directed search to smaller decision trees, recent attention to Monte Carlo Tree Search (MCTS) [5] has brought modern reinforcement learning strategies into the forefront [6]. Enzenberger and Muller [7] open-sourced a Go Engine based on Monte-Carlo Tree Search and achieved competitive results as measured by its 2290 Elo rating points in 19x19 Go playing against other state-of-the-art programs. SRI had a research team working on a DARPA funded Go seedling project built on the same Go MCTS engine called Fuego [7]. Google's DeepMind team defeated the 18-time world Go champion, Lee Sedol, using MCTS [8, 9]. Through a combination of playing many millions of Go games and directing its look-ahead (50-60 moves) reward search, the machine learning world has brought a new paradigm to solving high-dimensional games [10]. AlphaGo employed some novel strategies, such as previously ignored "slack" moves which to most champions appear unproductive, wasted moves. Observers such as the prominent venture capitalist, Kai-Fu Lee [11] have commented that Asia's embrace of machine learning curricula largely took off from this singular human-machine challenge, given that if an algorithm can win at Go, it can do other amazing things [12]. Pointedly, the Go defeat has been compared to a "Sputnik moment" for Asian countries and ignited further AI investments [11].

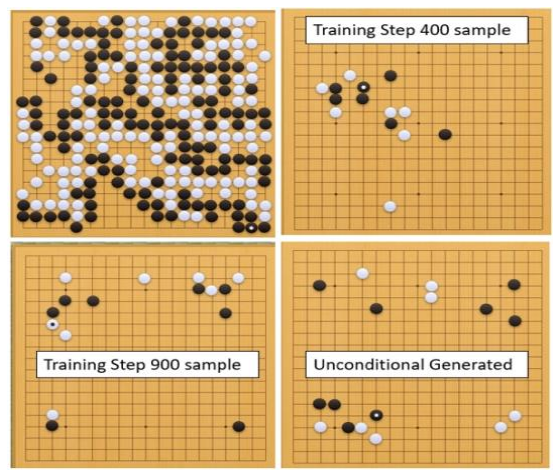

Fig. 1. Input Go game (actual upper left), transformer generated play after 400, 900 training steps and random inference post-training. The input game notable is white victory, Aizu Cup game, and a 317-move input game from the championship dataset.

A second important thread in the machine learning community has featured the rapid advances in natural language processing (NLP). Transformers or encoding of multiple languages into a truncated (but entirely numerical) search space offers a powerful strategy for understanding and enumerating multiple language tasks, including translation, summarization, topic modeling, and entity extraction [13]. For those who view language as the apex of all human skills compared to other species, the advances in computer language understanding seem profound. In a plausible and productive way, the ability to look forward and backwards in a text and guess the next sequences suggests that this transformer approach may yield interesting strategies outside of language alone. Image-text joint embeddings offer a recent example called ImageBERT [14], which derives a cross-modal relation between imagery and text much like other efforts to jointly predict new relations between audio, visual and language recognition skills.

Open AI's Generative Pre-trained Transformer, or GPT-2 [15] ,encodes multiple downstream language tasks with fine-tuned weights. One highly notable task follows from GPT-2's remarkable mimicry of plausible, coherent language generation [16]. The transformer provides a language model with 1.5 billion parameters. While trained on 8 million webpages, its goal reduces to predicting the next word in a text chunk, given all the previous words. Unlike other multi-task transformers, GPT-2 and its larger successors (GPT-3) have proven to be robust generators of synthetic text, particularly as a mimic of style and content after fine-tuning on an author's training text. Noteworthy applications of GPT-2 to play text adventure games like AI Dungeon and Choose Your Story have already included extensive machine-only game play and contests [17].

At first game play seems a challenging arena for a text generator: most games offer rigid rules and enumerated choices that make natural language processors an unlikely ally. Its rigidity (particularly the smaller 124M-GPT-2) propagates a strict and structured style that tends to memorize a game format as much as generalizing one. To generate player moves in the ancient Go game, we fine-tuned GPT-2 models using a large repository of professional games [18]. When play gets encoded in ASCII text, GPT-2 can emulate and generalize a professional's game style. The trained model further generates valid but previously unseen strategies for Go. Because GPT-2 preserves punctuation and spacing, the raw output of the text generator provides inputs to game visualization and creative patterns, such as the Sabaki project's [19] game engine using auto-replays. For simplicity sake, we call this the "Go Transformer" in the Results section, but it is more correctly a fine-tuned GPT-2 model that uses SGF text formats to learn the player moves in Go from historical data only. Remarkably, although GPT-2 was originally trained on highly ranked conversations from Reddit, it quickly adapts to mirror the Go game format and its restricted vocabulary. It is likely that no part of GPT-2 training would include any tokenized aspects that appear in most SGF games, which in shorthand encode player roles (black or white, B/W) and grid moves (two-letter row-column designations between lower letters "a" to "s" only).

## II. METHODS

### A. Datasets and Pre-processing

For fine-tuning GPT-2 from training data, we used the 56,638 Go Games from a championship repository [18]. Each game uses the Smart Game Format (SGF v.4), a UTF-8 file format common to many board games, including Go [20], Connect-4 and 6, and Amazons [21]. Not only does this text encoding annotate the game moves among the 361 intersections of a Go board, SGF offers no strict error-checking so that even a rough game can still be displayed in various visualization. In addition to metadata, the black (B) and white (W) stones are given two lower case letters (a-s) to signify the grid intersections on a 19x19 board. So a two-move sequence in SGF appears as ";B[qd];W[pp]", where the move delimiter is a semi-colon and Black starts always (in this case with row "q" and column "d". The SGF format also doesn't need a line break to delimit plays but for readability displays 10 moves per row. The SGF format offers a simple training set with an optional metadata header describing square bracketed tags for the event (EV), round (RO), player black name (PB) and rank (BR), player white name (PW) and rank (WR), time limit (TM, typically in hours), Komi (KM, or white's compensation points between 5-8 for going second), result (RE, in winning color and point advantage, e.g. W+13.5), and date (DT). The rest of the game alternates Black (B) and White (W) moves. The first objective of generating plausible Go games in SGF format therefore features the challenge to capture the syntax of alternating moves, with Black always the first player.

A reasonable game generates hundreds of moves, so if a single move is 6 characters (e.g, ";W[nd]") the expected text generated version here would need at least to generate on the order of 2000 characters or tokens. Since GPT-2 can only generate 1024 tokens per request (about 3-4 paragraphs), it currently doesn't lend itself to fine-tuning a complete game for Go unless the move sequences were short. For our purposes we want to evaluate the language transformers ability to specialize to SGF formats and its Go-related vocabulary to describe coherent moves and even playing formations or partial styles. In this way, the goal is not unlike previous work using transformers to encode poetry, music, film scripts, and question/answer pairs. by emulating their highly stylized vocabulary, stanza, and iconic page layouts. To preserve the coherence of a single game in SGF, we introduce standard text-preprocessing steps, stripping off all non-ASCII characters and removing line breaks in favor of a single pipe "|" delimiter, such that each game is treated during GPT-2 training as a single line. We have not experimented with "trigger" prompts or prefixes that would signal the start (<|startoftext|>) and end (<|endoftext|>) for each line, which can prove useful for the text generation as a kind of 'conversation starter' and overall delimiter. Initial experiments using the SGF start and end parenthetical tokens [(: to )] had the unforeseen effect of mixing a lot of out-of-game vocabulary and triggering a messier output with pre-trained inputs. A second objective therefore stems from the challenge for GPT-2 to understand the highly restricted Go-

SGF syntax and the game rhythm. For example, a less valuable opening move is an edge move because it neither offers much captured territory in the future and also restricts movement (or "liberties"). Similarly opening moves too close to the board center require long stretches of connected stones before walling in a board corner.

*B. Models and Train/Test Cycles*

The transformer selected for these experiments builds on the fine-tuning made simple by Woolf [22] python package, gpt-2-simple, which wraps OpenAI's Generative Pre-trained Transformer [15] for the "small" (124M), "medium" (355M) and "large" (744M) hyperparameter models. These models fit on most modern GPU cards. We initially trained and generated experimental Go games on laptop with NVIDIA Quadro M1200 with 4 GB of GDDR5 RAM and then scaled to 4x V100s. It is worth noting that the original pretrained 124 M model takes 500 MB, as does its fine-tuned ancestors here, and the pre-trained 355 M and 744M model is 1.5 GB and 3.2 GB respectively. Currently a modern (single V100) GPU can fine-tune the larger transformer (774M) models. We benchmarked 1000 training steps which takes around 13.5 hours on a M1200 GPU. We extended that model to 40,000 training steps on the V100, with the "medium" 355 M hyperparameter pretrained models (2.4 hours) and the "large" 744M model (11 hours). The cross-entropy log loss curve begins to flatten at 0.82, suggesting that to achieve better results (~0.25) we would benefit from a slower learning rate or larger batch size. Typically, our text sample for SGF games (56,638 games or 78 MB) is much larger than most transfer-learning examples and thus would need more training steps to sample all the input data consistently. If one compares the Go Game dataset to text alone, its approximately 16 million Go moves corresponds to a training text of comparable word count (average English word is 4.7 characters plus space token compared to average Go move in SGF as 5 letters plus semi-colon delimiter). In other words, we are training on nearly a 36,000-page corpus (100 books), in which every word is the same length, contains at least three common letters ("[];") and must start with only two characters (capital B or W).

We have not altered the default values for batch size (1), learning rate (0.0001), and 'Adam' optimizer, although larger GPU memory (>32 GB) could handle larger batch size and slower learning rates to outperform this fine-tuning example (Woolf, 2020). Every 100 steps we sample the checkpoint and generate a Go game of length 1024, or around 170 moves maximally. We generate unconditional samples randomly without user input or starter prompts. Most of the coherent opening play ranges from 30-50 moves.

## III. RESULTS

As shown in Figures 1,2 and 3, the Go transformer generates coherent player moves. The finetuning picks up on the alternating B/W stones and limits its output to realizable 19x19 grid points (a-s lowercase only). It mimics the standard delimiters of semicolons between moves, and square brackets between grid intersections. The transformer often doesn't generate reasonable start and end tokens in our training method, which may be a candidate for future improvement. Like the original GPT-2, some random generated samples prove less meaningful or lack a complete game cycle with a closed parenthesis and an opening (";") or header. Given SGF's lack of strictness in interpreting these game notes, the games play in auto-mode mode quite well in most cases.

We illustrate some example visualizations using the Sabaki [19] project in Figures 1-2. For comparison, we also highlight what a full 300+ move game looks like from training data (Fig.1) and generate random game play in Figure 2 shown next to the medium 355 M GPT-2 model finetuned after 11,000 training steps. Some notable evidence of training is seen between random and Go transformer play, since random play occupies poor starting positions, such as the board edges and second rows. This adverse start from random play doesn't arise from the Go transformer, which mimics the standard concepts for efficient opening, namely that corner play (Fig.2) bests side and center play for gaining territory.

As shown in, the fine-tuned GPT-2 models capture the basic game rhythm and notation. Black always plays first. Alternating player colors align (White and then Black) for the full game. The interior grid points have the required intersection letters without drifting off the board or repeating already played spaces. The overall coherence of play seems to follow the expected logical formations and sequencing as the championship training dataset. Figure 3 shows the result of medium GPT-2 finetuned for 41k training steps. The resulting model generates a nearly complete and coherent game sequence without any human knowledge, heuristic rules or strategic guidance. Like language transformers, the resulting exchange represents an end-to-end solution to game generation. This result contrasts markedly with other game approaches which inject reward and weighting functions to favor well-known or highly favorable moves. Compared to MCTS itself, the output does not obviously rely on look-ahead search or playing out forward scenarios to 50-60 moves to evaluate rewards. In most transformer architectures other than BERT [23], the previous move fires off the most probably next sequence to follow, rather than exploring all possible or locally deep forward moves.

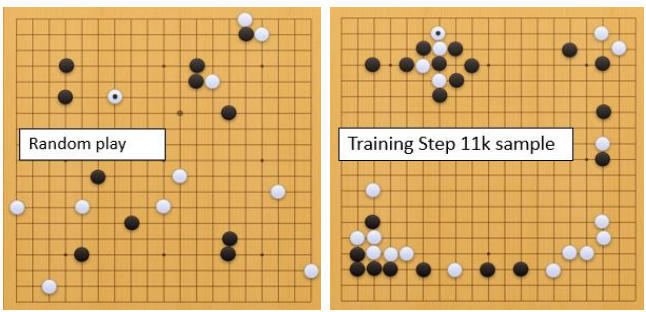

Fig 2. Trained 110K steps vs. Random play. Notable corner play and tighter stone formations contrast with random dispersions.

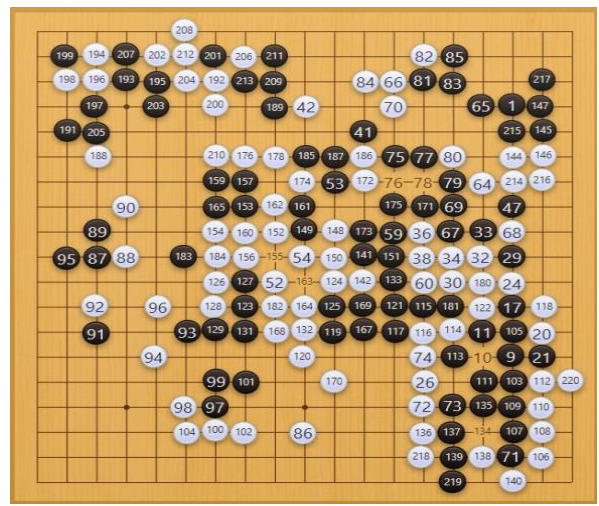

Fig 3. Medium GPT-2 Go Transformer play after 41k training steps. White leads in both Area and Territory, by 31 and 27 respectively. Number indicates play order.

IV. DISCUSSION

These results suggest that so-called universal language models, whether transformer-based or other deep learners, may offer novel ways to interact with other pattern-recognition and pattern-generating problems. One might conclude that compared to local search methods like Monte Carlo Tree Search, the complexity of sequences in language transformers offers an alternative method. We have crystallized the output of plausible and in some cases sophisticated Go formations and visualized the effects of two-player games. The Go transformer successfully encodes the largely esoteric tokens used by SGF historical archives and processes them in large quantities of greater than 50k gaming hours (approximately 6 years of championships playing 365/24/7). We anticipate future analysis to compare larger transformer models (>124 M hyperparameters), longer training times and alternate architectures (BERT, GPT-3, etc.). We also look forward to understanding better about the generated strategies in longer games with more Go formation plans.

TABLE I. TEXT FORMAT FOR GAMES

| Text Game Format | Supported Games with Possible Data |
| --- | --- |
| Smart Game Format (SGF) | Amazons, Ataxx, Backgammon, Byte, Chase, Chess, Dvonn, Exxit, Focus, Gess, Gipf, Go, Gobblet, Gomoku+Renju, Hex, Hive, Hnefatafl, Jungle, Kropki, Kuba, Lines of Action, Neutron, Nine Men's Morris, Octi, Othello, Philosopher's Football, Phutball, Plateau, PÜNCT, Quadrature, Sahara, Shogi, Tamsk, Tantrix, Trax, Tripples, Tumbling Down, TwixT, Xiangqi, Yinsh, Zèrtz, |
| Portable Game Notation | Chess |
| Portable Draughts Notation | Checkers, Draughts |
| Bridge Notation | Bridge |
| Video Games and Simulator Grammars | Multiple vintage games but SimCity 2000, Pirates, Minecraft, ZZT, etc. |

Since this work demonstrates for the first time, a simple transformer generating Smart Game Format, a fertile expansion of this problem includes other games supported by SGF. For example, Table 1 highlights future applications to more than 50 games beyond the current Go Transformer. This abbreviated list grows much larger if one imagines reverse engineering any number of game formats from a binary into hex code game files, which GPT-2 would mimic as two-digit numbers in sequence, then reassemble into a valid binary when successful.

ACKNOWLEDGMENTS

The authors would like to thank the PeopleTec Technical Fellows program for encouragement and project assistance. We are grateful to the front-line emergency workers who do their difficult work during the COVID-19 pandemic.

REFERENCES

[1] Singh, S., Okun, A., & Jackson, A. (2017). Artificial intelligence: Learning to play Go from scratch. Nature, 550(7676), 336-337.

[2] Bory, P. (2019). Deep new: The shifting narratives of artificial intelligence from Deep Blue to AlphaGo. Convergence, 25(4), 627-642.

[3] Fu, M. C. (2016, December). AlphaGo and Monte Carlo tree search: the simulation optimization perspective. In 2016 Winter Simulation Conference (WSC) (pp. 659-670). IEEE.

[4] Chen, J. X. (2016). The evolution of computing: AlphaGo. Computing in Science & Engineering, 18(4), 4-7.

[5] Fu, M. C. (2017). Markov decision processes, AlphaGo, and Monte Carlo tree search: Back to the future. In Leading Developments from INFORMS Communities (pp. 68-88). INFORMS.

[6] Lapan, M. (2018). Deep Reinforcement Learning Hands-On: Apply modern RL methods, with deep Q-networks, value iteration, policy gradients, TRPO, AlphaGo Zero and more. Packt Publishing Ltd.

[7] Enzenberger, M., Müller, M., Arneson, B., & Segal, R. (2010). Fuego—an open-source framework for board games and Go engine based on Monte Carlo tree search. IEEE Transactions on Computational Intelligence and AI in Games, 2(4), 259-270.

[8] Silver, D., Hubert, T., Schrittwieser, J., Antonoglou, I., Lai, M., Guez, A., ... & Lillicrap, T. (2017). Mastering chess and shogi by self-play with a general reinforcement learning algorithm. arXiv preprint arXiv:1712.01815.

[9] Silver, D., Hubert, T., Schrittwieser, J., Antonoglou, I., Lai, M., Guez, A., ... & Lillicrap, T. (2018). A general reinforcement learning algorithm that masters chess, shogi, and Go through self-play. Science, 362(6419), 1140-1144.

[10] Dong, X., Wu, J., & Zhou, L. (2017). Demystifying AlphaGo zero as AlphaGo GAN. arXiv preprint arXiv:1711.09091.

[11] Lee, Kai-Fu, (2018) "China's 'Sputnik Moment' and the Sino-American Battle for AI Supremacy", Asia Society, https://asiasociety.org/blog/asia/chinas-sputnik-moment-and-sino-american-battle-ai-supremacy

[12] Wang, F. Y., Zhang, J. J., Zheng, X., Wang, X., Yuan, Y., Dai, X., ... & Yang, L. (2016). Where does AlphaGo go: From church-Turing thesis to AlphaGo thesis and beyond. IEEE/CAA Journal of Automatica Sinica, 3(2), 113-120.

[13] Klein, T., & Nabi, M. (2019). Learning to Answer by Learning to Ask: Getting the Best of GPT-2 and BERT Worlds. arXiv preprint arXiv:1911.02365.

[14] Qi, D., Su, L., Song, J., Cui, E., Bharti, T., & Sacheti, A. (2020). ImageBERT: Cross-modal pre-training with large-scale weak-supervised image-text data. arXiv preprint arXiv:2001.07966.

[15] Radford, A., Wu, J., Child, R., Luan, D., Amodei, D., & Sutskever, I. (2019). Language models are unsupervised multitask learners. OpenAI Blog, 1(8), 9. https://github.com/openai/gpt-2

[16] Vig, J. (2019). OpenAI GPT-2: Understanding Language Generation through Visualization. Towards Data Science, via Medium, March, 5.

[17] Kirubarajan, A., & Dugan, L. (2020) Learning to Trick Humans: Evaluation Criteria for Human-Written and Computer-Generated Text.

[18] Go Games, https://homepages.cwi.nl/~aeb/go/games/index.html

[19] Sabaki Project, (2020) An elegant Go Board and SGF editor for a more civilized age. https://github.com/SabakiHQ/Sabaki

[20] Kierulf, A. (1987, December). Human-computer interaction in the game of Go. In Proceedings of the Second International Symposium on Methodologies for intelligent systems (pp. 481-487).

[21] Müller, M., & Tegos, T. (2002). Experiments in computer Amazons. More Games of No Chance, 42, 243-260.

[22] Woolf, Max, (2020), GPT-2-Simple, a Python Package, https://github.com/minimaxir/gpt-2-simple

[23] Devlin, Jacob, et al. "Bert: Pre-training of deep bidirectional transformers for language understanding." arXiv preprint arXiv:1810.04805 (2018)